\documentclass[pdflatex,sn-vancouver-num]{sn-jnl}

\usepackage[T1]{fontenc}
\usepackage[english]{babel}
\usepackage{comment}
\usepackage{xcolor}
\usepackage{amsmath,amssymb,amsfonts}
\usepackage{mathtools}
\usepackage{siunitx}
\usepackage{graphicx}
\usepackage{booktabs}
\usepackage{float}
\usepackage{algorithm}
\usepackage{algpseudocode}
\usepackage{bookmark}

\hypersetup{hypertexnames=false}

\makeatletter
\renewcommand{\@maketitle}{\newpage\null%
    \if@remarkboxon\vbox to 0pt{\vspace*{-78pt}\hspace*{-18pt}\FMremark}\fi%
    \hsize\textwidth\parindent0pt%
    {\hbox to \textwidth{{\Artcatfont\ArtType\hfill}\par}}%
    \ifx\@title\empty\else%
        \removelastskip\vskip4pt\nointerlineskip%
        {\Titlefont\@title\par}%
    \fi%
    \ifx\@subtitle\empty\else%
        \vskip6pt%
        {{\SubTitlefont\@subtitle\par}}%
    \fi%
    \ifnum\aucount>0
        \global\punctcount\aucount%
        \vskip10pt%
        \artauthors\par%
        {\vskip4pt\addressfont\auaddress\par%
         \removelastskip\vskip8pt%
        \ifnum\emailcnt>0\relax%
           \ifx\corrauthemail\@empty\else{\ifnum\aucount>1*\fi}%
           Corresponding author(s). E-mail(s): \corrauthemail\par\fi%
           \ifx\authemail\@empty\else Contributing authors:\ \authemail\fi%
        \fi%
        \ifequalcont{\par$^{\dagger}$\@equalconttext\par}\fi%
         \removelastskip\vskip8pt%
        \ifpresentaddress{\par\@presentaddresstext\par}\fi%
        }%
     \fi%
     {\printabstract\par}%
     {\printkeywords\par}%
     \ifx\@pacs\empty\else%
       \loop\ifnum\PacsCount>0%
          \csname\romannumeral\PacsTmpCnt StorePacsTxt\endcsname\par%
          \StepDownCounter{\PacsCount}%
          \StepUpCounter{\PacsTmpCnt}%
       \repeat%
    \fi%
    \removelastskip\vskip12pt\vskip0pt}%
\makeatother

\title[Compact Robot for Endovascular Interventions]{A Compact Top-Loading Robot for Endovascular Interventions: Design, Control and Evaluation}

\author[1]{\fnm{Jonas} \sur{Fischer}}\email{jonas.f.fischer@fau.de}
\equalcont{These authors contributed equally to this work.}

\author[1]{\fnm{Lennart} \sur{Karstensen}}\email{lennart@karstensen.biz}
\equalcont{These authors contributed equally to this work.}

\author*[1]{\fnm{Franziska} \sur{Mathis-Ullrich}}\email{franziska.mathis-ullrich@fau.de}

\affil*[1]{\orgdiv{Laboratory for Surgical Planning and Robot Cognition (SPARC) },
\orgname{Friedrich-Alexander-University Erlangen-N\"urnberg},
\orgaddress{\street{N\"urnberger Stra\ss e 74}, \city{Erlangen},
\postcode{91052}, \state{Bavaria}, \country{Germany}}}

\begin{document}

\abstract{
\textbf{Purpose:}
Robot-assisted endovascular intervention has the potential to reduce radiation exposure, improve surgeon ergonomics, enable telesurgery, support active assistance and autonomy, and enhance procedural precision. However, existing systems often suffer from limited procedural coverage because constrained patient-side setups, restricted flexibility, and complex instrument exchange hinder clinical workflow integration. This work presents a compact robotic system for endovascular interventions that enables continuous translational and rotational manipulation of standard endovascular instruments.

\textbf{Methods:}
The system consists of two alternating carts with pneumatically actuated membrane grippers integrated into rotating gripper gears. Its top-loading design allows rapid exchange of instruments such as guidewires and catheters without changing the robotic setup. A leader-follower control strategy enables continuous motion despite the finite stroke of each cart. The system was evaluated in motion-tracking experiments with guidewires and catheters and in an in vitro vascular phantom.

\textbf{Results:}
The motion-tracking experiments showed predominantly smooth translational and rotational motion profiles. Across all tested guidewire and catheter experiments, the mean relative tracking errors were \SI{3.6 \pm 2.2}{\percent} for translational motion and \SI{4.1 \pm 1.6}{\percent} for rotational motion. In the vascular phantom, robot-assisted navigation reached the target in most trials, demonstrating the feasibility of the proposed manipulation concept under in vitro conditions.

\textbf{Conclusion:}
The presented robotic system demonstrates technical feasibility for continuous manipulation of standard endovascular instruments in bench-top and in vitro experiments. The compact top-loading design may facilitate instrument exchange and clinical workflow integration. Future work will focus on improving gripping performance, actuation speed, force feedback, and evaluation in more clinically realistic settings.
}

\keywords{Robot-assisted endovascular intervention, Endovascular robotics, Robotic catheterization, Surgical robotics}

\maketitle

\section{Introduction}

\begin{figure}[htbp]
    \centering
    
    \includegraphics[width=1\linewidth, trim=5.4cm 3cm 0cm 3cm,
    clip]{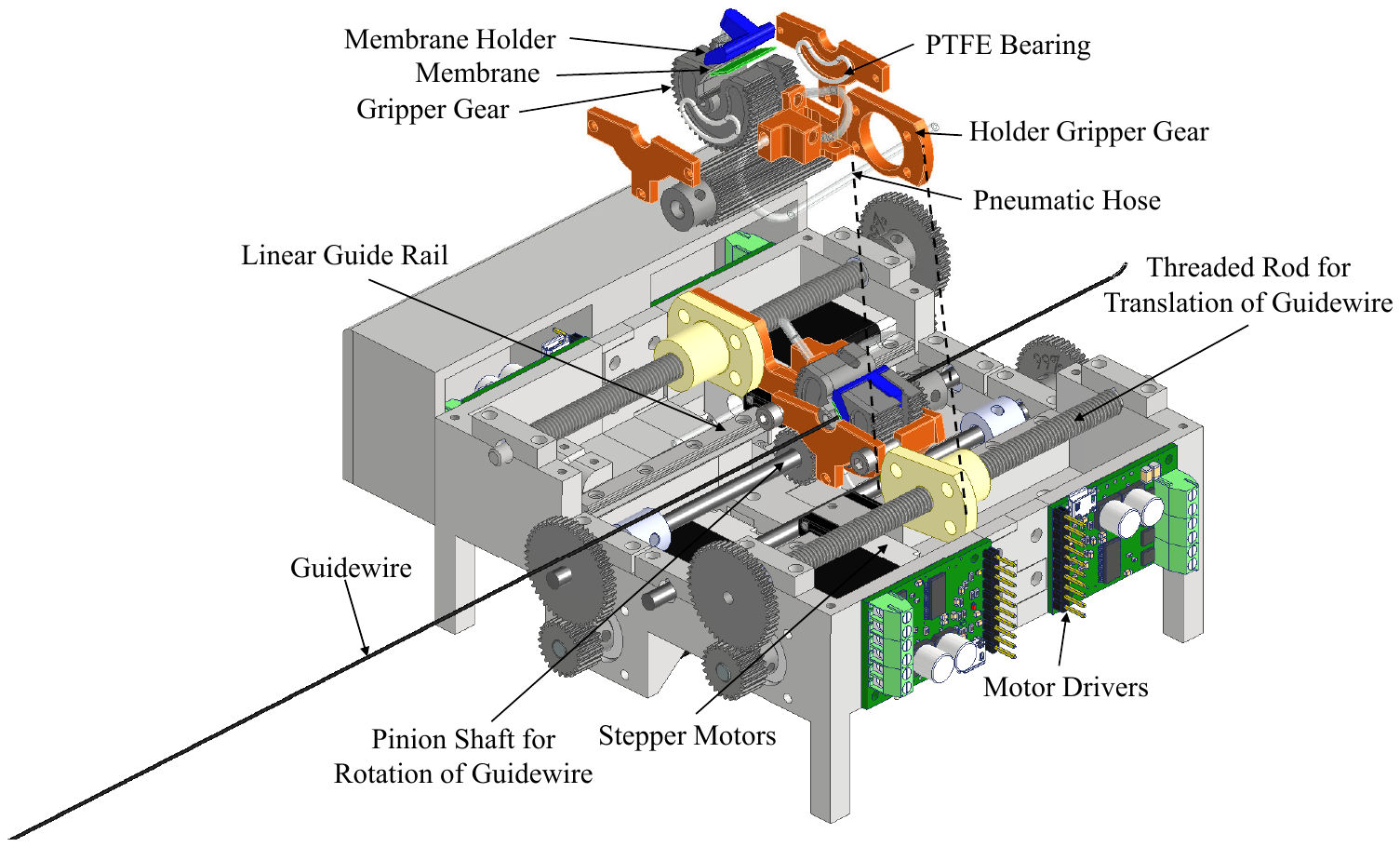}    
    \vspace{0.5em}
    \caption{CAD model of roboEVI with two carts, one in position and one in exploded view
}
    \label{fig:roboEVI_CAD}
\end{figure}

Cardiovascular and cerebrovascular diseases remain among the leading causes of morbidity and mortality worldwide, motivating continued development of safe and effective minimally invasive treatments \cite{pescio_endovascular_2025, al-ahmad_force_2023}. Endovascular interventions are a standard therapy for many vascular pathologies because they reduce surgical trauma, recovery time, and infection risk compared to open surgery \cite{pescio_endovascular_2025, xie2025design}. In these procedures, catheters and guidewires are advanced through the vasculature to a target region for diagnostic or therapeutic steps such as angioplasty, stenting, embolization, or ablation \cite{pescio_endovascular_2025}. Despite their clinical success, they remain demanding because operators must navigate thin, flexible instruments through delicate vessels while avoiding excessive contact forces that can cause injury or perforation \cite{pescio_endovascular_2025}.

Conventional endovascular navigation is typically performed under X-ray fluoroscopy. Although fluoroscopy provides real-time visualization, it exposes patients and clinical staff to ionizing radiation \cite{pescio_endovascular_2025,belikov_evolution_2024}. In addition, contrast-agent injections are required to visualize the vasculature and may be associated with nephrotoxicity and other complications \cite{pescio_endovascular_2025}. The occupational burden for physicians is also considerable because protective lead garments are heavy and prolonged standing can contribute to musculoskeletal problems \cite{belikov_evolution_2024, bao2018operation}. These limitations have motivated robotic systems for remote instrument manipulation to reduce radiation exposure, improve ergonomics, and increase motion precision \cite{pescio_endovascular_2025, belikov_evolution_2024}.

Commercial and research platforms have demonstrated the potential of robot-assisted endovascular interventions. Commercial systems such as Niobe, Sensei, Magellan, CorPath, R-One, Amigo, and Sentante illustrate different approaches to remote catheter or guidewire manipulation, including magnetic actuation, electromechanical drive units, joystick-based control, and haptic interfaces \cite{pescio_endovascular_2025}. Belikov et al. \cite{belikov_evolution_2024} reported that most follower-side robotic catheter systems provide two to three degrees of freedom and realize translation and rotation using rollers, belts, grippers, or magnetic actuation \cite{belikov_evolution_2024}. These systems can achieve clinically relevant motion ranges and accuracies in bench-top evaluations \cite{belikov_evolution_2024, bao2016design}, but important barriers to clinical adoption remain \cite{belikov_evolution_2024}.

Recent prototypes illustrate different approaches to these barriers, but also show that workflow integration, instrument exchange, patient-side compactness, and compatibility with standard clinical instruments remain key design challenges. Al-Ahmad et al. \cite{al-ahmad_force_2023} presented a robot-assisted catheterization system based on braided sleeve grippers for teleoperated deployment of generic endovascular instruments. It provides translation and rotation through linear-motor translation and spur-gear rotation of the gripper assembly, with synchronized dynamic regripping for continuous motion. Dagnino et al. \cite{dagnino_-vivo_2023} developed CathBot, a teleoperated platform with a leader device, an MR-safe remote manipulator, and pneumatic actuation. It mimics manual catheter and guidewire handling through push-pull and rotary inputs, while plug-and-play mechanisms enable quick exchange of standard vascular instruments, although setup and positioning still contributed to longer procedure times \cite{dagnino_-vivo_2023}. Song et al. \cite{song_novel_2023} evaluated the ALLVAS system for robot-assisted endovascular aortic repair. Its patient-side unit uses two independent mechanical arms with manipulators and grippers for advancing, retracting, rotating, and deploying guidewires, catheters, balloons, and stent grafts. Although compatible with rapid-exchange and over-the-wire devices, the depicted setup indicates a comparatively large patient-side footprint \cite{song_novel_2023}. Bao et al. \cite{bao2018operation} developed RobEnt, which uses a simplified two-unit structure with static grasping and a conical clamping principle to improve motion transmission while reducing slippage and surface damage \cite{bao2018operation}.

Taken together, these systems demonstrate important progress toward safer and more precise robotic endovascular intervention, but no single design has fully resolved the combined requirements of compactness, simple clinical handling, rapid instrument exchange, and robust compatibility with standard tools. In line with these challenges, Pescio et al. \cite{pescio_endovascular_2025} emphasize that future systems should improve compatibility with off-the-shelf equipment, interchangeability, navigation technologies, and workflow integration. Similar concerns appear in research prototypes: Bao et al. \cite{bao2018operation} noted that complex robotic structures can make catheter or guidewire installation and removal inconvenient, increase control complexity, and reduce operation efficiency, while Chen et al. highlight limited compatibility with different endovascular instruments as a limitation of current systems \cite{chen_remote_2026}. These observations indicate a research gap for compact robotic systems that simplify insertion and exchange of standard instruments while reducing dependence on complex assembly steps.

To address these workflow- and mechanism-related gaps, this work presents a compact robot for endovascular interventions. The system uses two alternating carts that translate and rotate the instrument. Each cart incorporates a pneumatically actuated membrane gripper integrated into a rotating gripper gear, allowing insertion from above. This top-loading concept enables rapid exchange of instruments such as guidewires and catheters without changing the robotic setup, and the alternating leader-follower strategy enables continuous motion despite the finite stroke of each cart. By combining compact dimensions, modular construction, pneumatic gripping, and rapid instrument exchange, the design addresses workflow-relevant requirements while enabling controlled manipulation of standard endovascular instruments.

\section{Methods}

\subsection{Hardware and Design}

The robot consists of two carts, as shown in Fig.~\ref{fig:roboEVI_CAD}. One cart is depicted assembled and the other in an exploded view. Each cart is actuated by two NEMA 8 stepper motors (Nanotec Electronic GmbH \& Co. KG, Feldkirchen, Germany), one for translation and one for rotation, controlled using Pololu TIC 36V4 drivers (Pololu Corporation, Las Vegas, USA). The translational motor drives the threaded rod, moving the cart through the gripper-gear holder (orange), while a linear guide rail guides the cart.

Rotational motion of the gripper gear is generated by the second stepper motor through a pinion shaft beneath the cart. The gripper gear is mounted in the gripper-gear holder using two polytetrafluoroethylene (PTFE) bearings on each side, and a thin layer of lubricating grease reduces friction and ensures smooth rotation. The endovascular instrument is inserted from above through the gap between the membrane (green) and the gripper gear. This top-loading design enables rapid exchange of instruments with different diameters, such as switching from a guidewire to a catheter. The instrument is grasped by a pneumatically actuated membrane inside the gripper gear and glued to the membrane holder (blue). When the membrane is pressurized to \SI{1}{bar} using the solenoid valve (MHA2-M1H-3/2O-2-K, Festo SE \& Co. KG, Esslingen am Neckar, Germany), it expands and presses the instrument against the rear surface of the gripper gear. The solenoid valves are controlled via a dual MOSFET trigger module (MakerMind, China). An Arduino Nano RP2040 microcontroller (Qualcomm, San Diego, USA) controls the robot and is powered by an AC/DC power supply (MEAN WELL Enterprises Co., Ltd., New Taipei City, Taiwan). The casing is CNC-machined, whereas the side and top covers, gears, and membrane holders are fabricated from PLA using a Bambu Lab X1 3D printer (Bambu Lab, Shenzhen, China). The gripper gear is fabricated by stereolithography (SLA) on a Form 4 printer (Formlabs Inc., Somerville, MA, USA), because its higher dimensional accuracy and tighter tolerances are required to ensure smooth rotation of the PTFE bearing.
The robot is designed to be compact, with overall dimensions of \SI{176}{\milli\meter} in width, \SI{137.6}{\milli\meter} in length, and \SI{63.2}{\milli\meter} in height.

\subsection{Control Algorithm}

\textbf{Alternating Grip Drive}

\begin{algorithm}
\caption{Leader/Follower Velocity Control}
\label{alg:set-velocities}
\begin{algorithmic}[1]
\Require Commanded velocities $v_t$ [mm\,s$^{-1}$], $v_r$ [deg\,s$^{-1}$]
\Require Leader cart $L$, follower cart $F$; limits $\bar{v}_t$, $\bar{v}_r$
\Ensure Clamp states and velocity commands applied to both carts

\State $v_t \leftarrow \mathrm{clamp}(v_t,\,-\bar{v}_t,\,\bar{v}_t)$
\State $v_r \leftarrow \mathrm{clamp}(v_r,\,-\bar{v}_r,\,\bar{v}_r)$

\If{$L.\textsc{StrokeEndReached}(v_t, v_r)$ \textbf{and} $F.\textsc{IsClamped}()$}
    \State $L \leftrightarrow F$ \Comment{swap leader and follower roles}
\EndIf

\State $L.\textsc{Clamp}()$
\State $L.\textsc{SetVelocities}(v_t,\, v_r)$

\If{$L.\textsc{StrokeEndReached}(v_t, v_r)$} \Comment{handover phase}
    \State $F.\textsc{SetVelocities}(v_t,\, v_r)$
    \If{$F.\textsc{VelocityReached}(v_t, v_r)$}
        \State $F.\textsc{Clamp}()$ \Comment{follower ready to become leader}
    \Else
        \State $F.\textsc{Release}()$
    \EndIf
\Else \Comment{normal phase}
    \State $F.\textsc{Release}()$
    \If{$F.\textsc{IsReleased}()$}
        \State $F.\textsc{SetTargetPosition}(0,\, 0)$ \Comment{return follower to home}
    \Else
        \State $F.\textsc{SetVelocities}(v_t,\, v_r)$ \Comment{coast during release delay}
    \EndIf
\EndIf
\end{algorithmic}
\end{algorithm}

Algorithm~\ref{alg:set-velocities} implements the core motion strategy. At any instant, one cart is the \emph{leader} and the other the \emph{follower}:
\begin{itemize}
  \item The \textbf{leader} is clamped and driven at the commanded velocity.
  \item The \textbf{follower} is released and returns to home position~$(0,0)$.
\end{itemize}

When the leader reaches the end of its usable stroke and the follower has already clamped, the roles are exchanged. The new leader immediately continues driving, while the old leader releases and homes. This cycle repeats without interrupting catheter velocity.

A short \emph{handover phase} bridges the transition: when the leader reaches its stroke limit, the follower is commanded to match the target velocity before clamping. This ensures the correct speed at the moment of swap and prevents a velocity transient from being transmitted to the catheter.
\newpage
\noindent\textbf{Maximum Velocities}

The following quantities are used to estimate the maximum continuous instrument velocity:
\begin{itemize}
\item $V_{max}$: maximum actuation speed of the instruments
\item $L_{S/2}$: half of the stroke length of one actuator
\item $V_{ret_{max}}$: absolute maximum speed of an actor during return motion
\item $t_{clamp}$: time required to fully clamp the instrument
\item $L_{clamp_{max}} = t_{clamp} \cdot V_{max}$: maximum length required for clamping the instrument
\item $t_{ret} = L_{S/2} / V_{ret_{max}}$: maximum time required to return the cart from the stroke end to the home position
\end{itemize}

The usable drive length in each direction must satisfy:
\begin{equation}
L_{D/2} \leq L_{S/2} - 2 \cdot L_{clamp_{max}}
\end{equation}

while return to home requires:

\begin{equation}
    L_{D/2} \geq 2 \cdot L_{clamp_{max}} + V_{max} \cdot t_{ret}
\end{equation}

Therefore

\begin{equation}
    L_{S/2} - 2 \cdot L_{clamp_{max}} \geq 2 \cdot L_{clamp_{max}} + V_{max} \cdot t_{ret}
\end{equation}
\begin{equation}
    L_{S/2} \geq 4 \cdot t_{clamp} \cdot V_{max} + V_{max} \cdot t_{ret}
\end{equation}
\begin{equation}
     \frac{L_{S/2}}{4 \cdot t_{clamp} + \frac{L_{S/2}}{V_{ret_{max}}}} \geq V_{max}
\end{equation}

For the implemented prototype, the feasible translational stroke is set to \(L_{S,\mathrm{trans}}=\SI{17.5}{\milli\meter}\), resulting in an available half-stroke of \(L_{S/2,\mathrm{trans}}=\SI{8.75}{\milli\meter}\). Using the clamping time \(t_{\mathrm{clamp},\mathrm{trans}}=\SI{50}{\milli\second}\) and the maximum return velocity \(V_{\mathrm{ret}_{\max},\mathrm{trans}}=\SI{100}{\milli\meter\per\second}\), the theoretical maximum translational instrument velocity is \(V_{\max,\mathrm{trans}}=\SI{30.4}{\milli\meter\per\second}\).

For rotational motion, the same analysis is applied with a feasible stroke of \(L_{S,\mathrm{rot}}=\SI{78.4}{\degree}\), corresponding to \(L_{S/2,\mathrm{rot}}=\SI{39.2}{\degree}\). With \(t_{\mathrm{clamp},\mathrm{rot}}=\SI{50}{\milli\second}\) and a maximum rotational return velocity of \(V_{\mathrm{ret}_{\max},\mathrm{rot}}=\SI{180}{\degree\per\second}\), the theoretical maximum angular instrument velocity is \(V_{\max,\mathrm{rot}}=\SI{93.8}{\degree\per\second}\).

\section{Experimental setup}

\begin{figure*}[!t]
    \centering
    
    \includegraphics[width=\linewidth, trim=0cm 5cm 0 5cm]{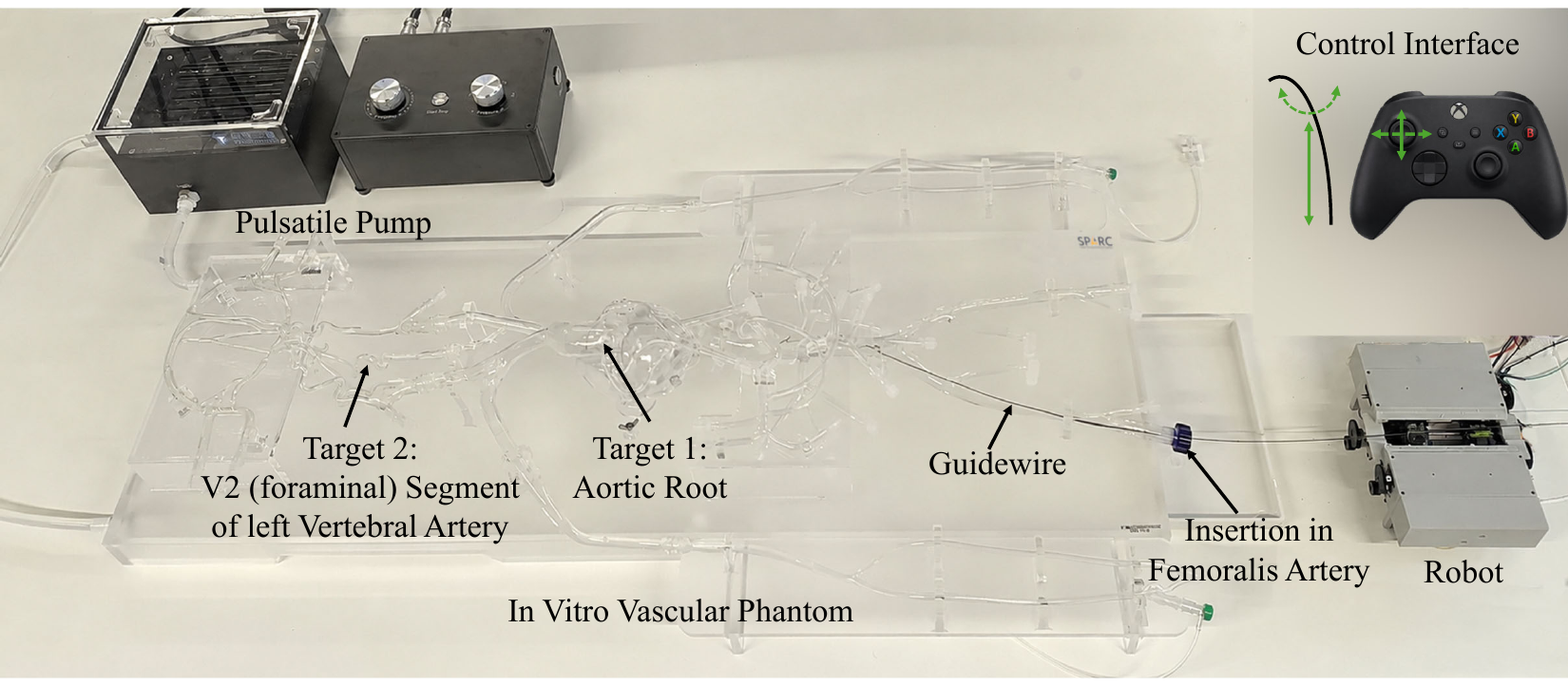}

    \vspace{0.5em}

    \caption{In vitro vascular-phantom setup of the robot with pulsatile flow and gamepad control. A guidewire is inserted through the femoral access site and navigated toward the aortic root (Target 1) and left vertebral artery V2 segment (Target 2)}
    
    \label{fig:roboEVI_Phantom}
\end{figure*}

To assess manipulation accuracy and usability, a series of experiments was conducted. First, manipulation accuracy was assessed in translational and rotational directions. Subsequently, navigation experiments were performed in a vascular phantom to evaluate the system in a more realistic clinical setting. The selection and design followed the recommendations of Robertshaw et al. \cite{robertshaw2026position}, derived from a Delphi study, including realistic vascular anatomy, deformable vessels, simulated blood flow and pulsatility, catheters and guidewires, and outcome measures such as success rate, number of unintended branch insertions, and procedure time.

\subsection{Robot Motion Tracking Evaluation }

Four experiments were conducted to assess tracking performance for different motion patterns and instruments. The evaluated instruments were a standard guidewire (\SI{0.035}{\text{in}}, stiff, Terumo, Tokyo, Japan) and a catheter (\SI{5.2}{\text{Fr}}, Cordis, Miami Lakes, USA). Each instrument was assessed for translational and rotational motion at different velocities.

\begin{figure}[htbp]
    \centering
    
    \includegraphics[width=\linewidth, trim={8cm 3cm 8cm 3cm}]{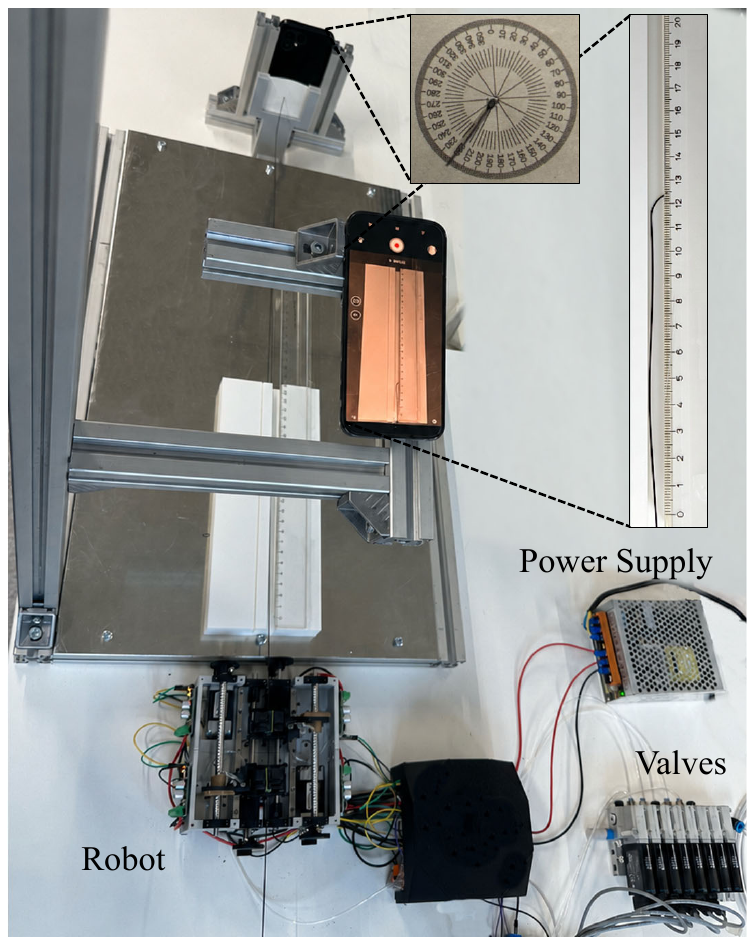}

    \vspace{0.5em}

    \caption{Motion-tracking setup of the robot with the actuation unit, power supply, pneumatic valves, and camera-based angular and linear reference scales}
    
    \label{fig:exp_setup}
\end{figure}

Fig.~\ref{fig:exp_setup} illustrates the setup. The robot, power supply, and pneumatic valves are located at the bottom. The robot translates the instruments along a white 3D-printed groove, while motion is recorded from above. Rotational movement is recorded from the front. Rotational displacement is measured by passing the proximal straight end of the instrument through a paper disk with angular graduations and using a small attached flag as an angle indicator.

All experiments were recorded using a Fairphone 5 smartphone camera (Fairphone, Netherlands) in slow-motion mode at \SI{240}{fps}. To extract positional information, k-nearest neighbor (kNN) models were trained separately for each instrument and velocity condition. For each translational run, \num{200} images evenly distributed across the \SI{200}{\milli\meter} travel distance were manually labeled, whereas \num{300} images were labeled for each rotational run. For translational analysis, model hyperparameters, including crop region, feature resolution, number of neighbors, and temporal smoothing window, were selected by validation on the labeled image set. These datasets served as training data, with image pixel values as input features and predicted position as output. The average mean prediction error was \SI{0.96 \pm 1.0}{\milli\meter} for translational models and \SI{1.1 \pm 0.2}{\degree} for rotational models.

For translational experiments, velocities of \SI{\pm5}{\milli\meter\per\second}, \SI{\pm10}{\milli\meter\per\second}, \SI{\pm15}{\milli\meter\per\second}, and \SI{\pm19}{\milli\meter\per\second} were investigated, spanning the practically feasible maximum translational range of \SI{\pm19}{\milli\meter\per\second}. Although the theoretical handover-limited velocity is \SI{30.4}{\milli\meter\per\second}, the maximum reliably achievable velocity in the implemented setup is \SI{19}{\milli\meter\per\second}, as determined in initial experiments. Each instrument was translated over \SI{200}{\milli\meter} at every tested velocity.

For rotational experiments, angular velocities of \SI{\pm5}{\degree\per\second}, \SI{\pm20}{\degree\per\second}, \SI{\pm35}{\degree\per\second}, and \SI{\pm52}{\degree\per\second} were investigated, spanning the practically feasible maximum rotational range of \SI{\pm52}{\degree\per\second}. Although the theoretical handover-limited angular velocity is \SI{93.8}{\degree\per\second}, the maximum reliably achievable angular velocity in the implemented setup is \SI{52}{\degree\per\second}, as determined in initial experiments. Each instrument was rotated by \SI{360}{\degree} at every tested angular velocity.

\subsection{In Vitro Phantom Experiment}

While the motion-tracking evaluation provides insight into manipulation accuracy, in vitro phantom experiments were conducted under conditions closer to clinical practice. Fig.~\ref{fig:roboEVI_Phantom} shows the setup, consisting of a pulsatile pump mimicking the heartbeat, an in vitro vascular phantom (Trandomed, Ningbo, China), the robot, and an Xbox controller (Microsoft, USA) with the corresponding control scheme. Moving the left joystick upward or downward corresponded to guidewire insertion or retraction, whereas moving it left or right resulted in positive or negative rotation.
The phantom was filled with a water-glycerin mixture of 35\% glycerin and 65\% water to reproduce blood viscosity \cite{yousif_blood-mimicking_2011}. 
Four experimental modes were defined based on target location and actuation method: \textit{aortic root manual}, \textit{aortic root robotic}, \textit{vertebral artery manual}, and \textit{vertebral artery robotic}. The targets were the aortic root and the V2 foraminal segment of the left vertebral artery, as marked in Fig.~\ref{fig:roboEVI_Phantom}, representing different levels of navigational complexity in vessel geometry and bifurcations. Navigation to the vertebral artery was considered more challenging because of the smaller vessel diameters and greater number of bifurcations along the access path. The starting point was always the femoral artery. Each mode was repeated five times. All experiments were conducted by a non-clinical operator experienced with the robotic system.

Evaluation metrics were success rate, unintended branch insertions, and mean time to reach the target. Success was defined as reaching the target within \SI{2}{\minute}, a limit determined during initial trials. Unintended branch insertions were defined as insertions of the guidewire into an incorrect vessel branch.

\section{Results}
\subsection{Robot Motion Tracking Evaluation }

\begin{figure}[htbp]
    \centering
    \includegraphics[
        width=\linewidth,
        trim={0cm 4cm 0cm 4cm},
        clip
    ]{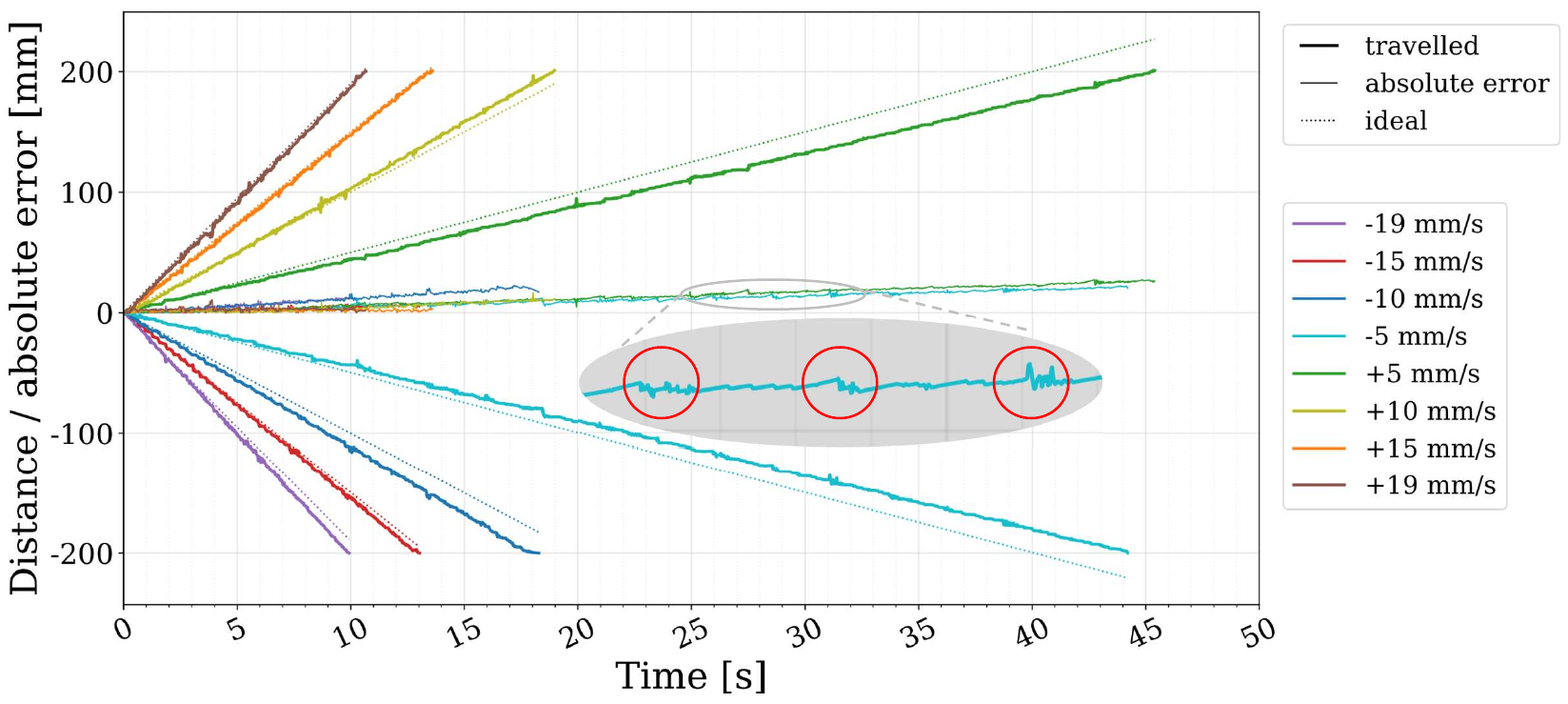}
    \caption{
        Guidewire translation at insertion and retraction velocities. Solid lines show measured motion, dashed lines ideal profiles, and thin lines absolute error. Red circles indicate leader--follower switching
    }
    \label{fig:gw_trans}
\end{figure}

\begin{figure}[htbp]
    \centering
    \includegraphics[
        width=\linewidth,
        trim={0cm 4cm 0cm 4cm},
        clip
    ]{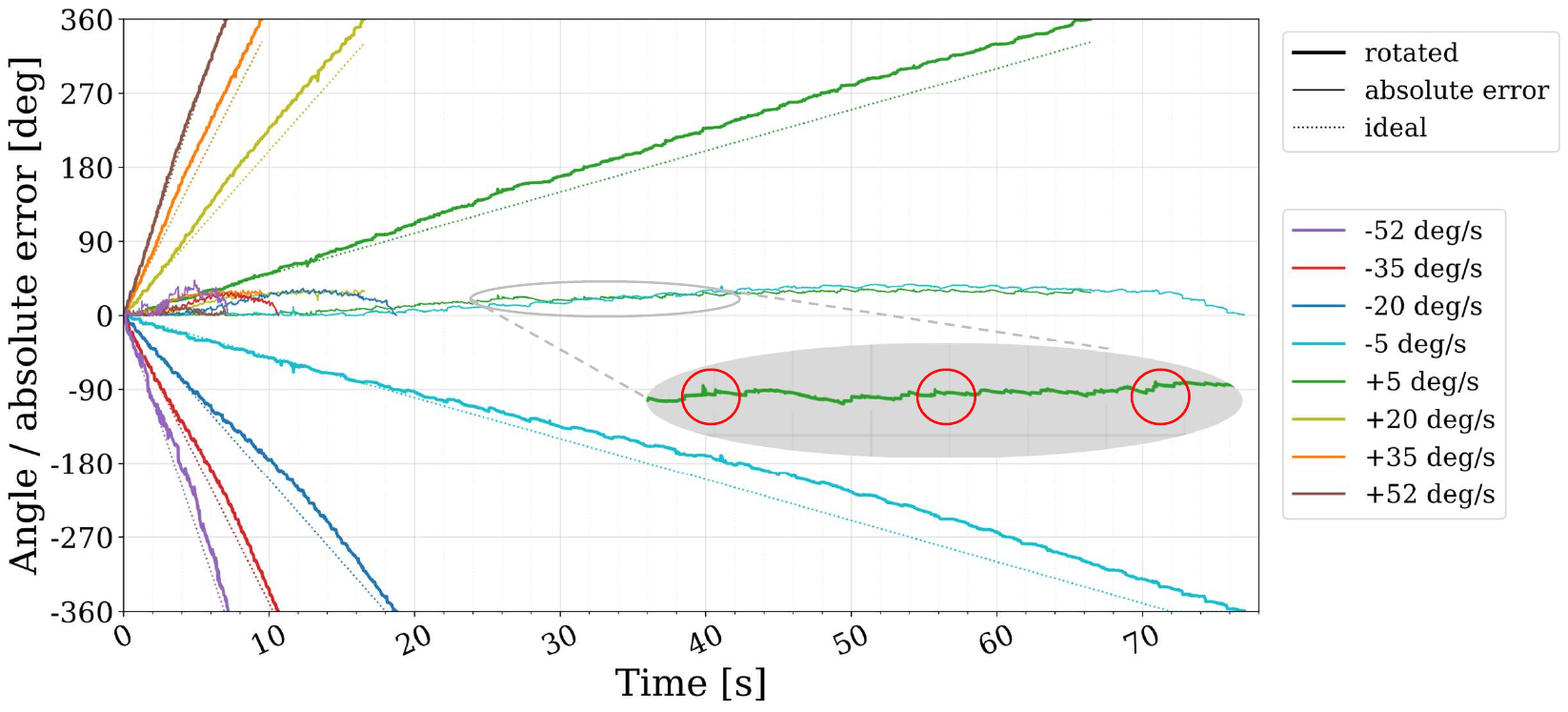}
    \caption{
        Guidewire rotation at tested positive and negative angular velocities. Solid lines show measured rotations, dashed lines ideal profiles, and thin lines absolute error. Red circles indicate leader--follower switching
    }
    \label{fig:gw_rot}
\end{figure}

\begin{figure}[htbp]
    \centering
    \includegraphics[
        width=\linewidth,
        trim={0cm 4cm 0cm 4cm},
        clip
    ]{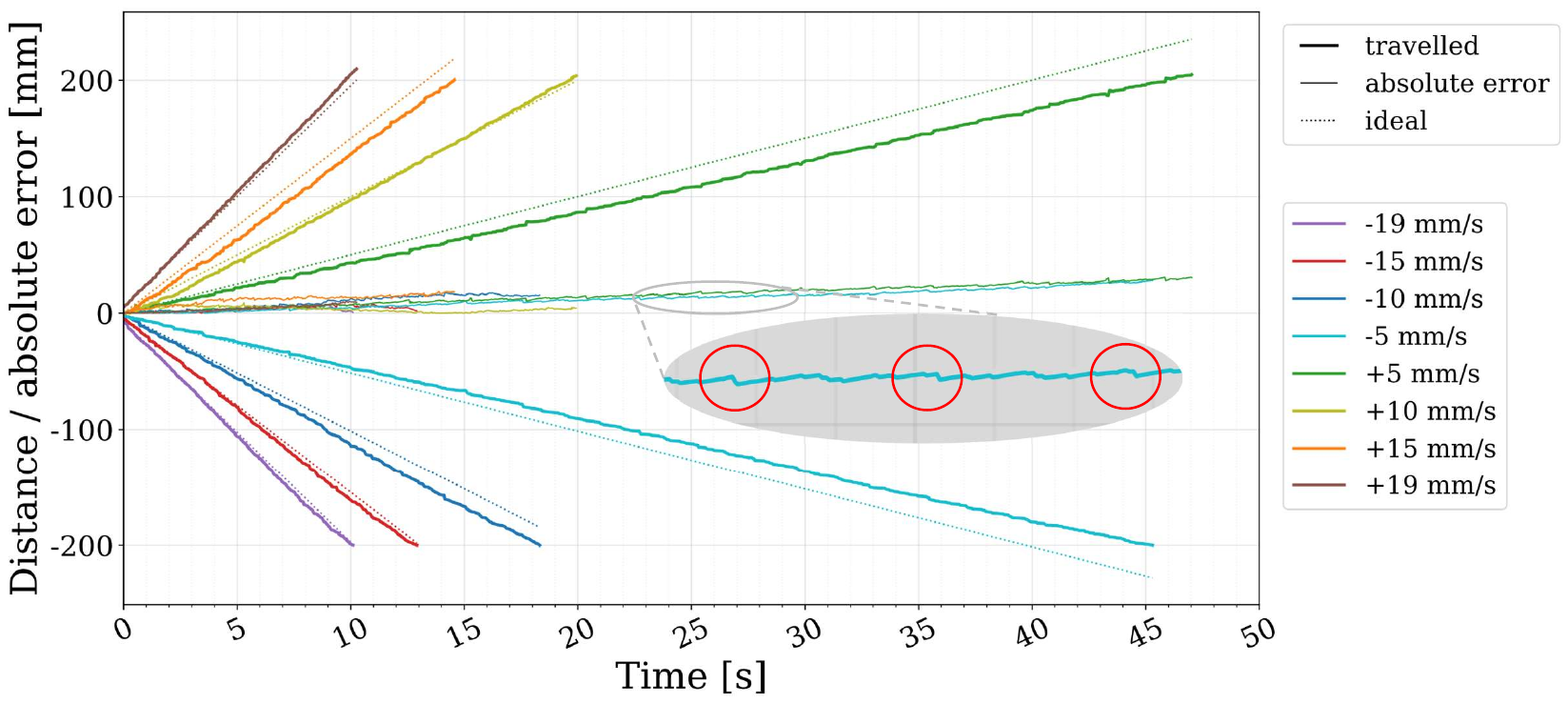}
    \caption{
        Catheter translation at insertion and retraction velocities. Solid lines show measured motion, dashed lines ideal profiles, and thin lines absolute error. Red circles indicate leader--follower switching
    }
    \label{fig:cath_trans}
\end{figure}

\begin{figure}[htbp]
    \centering
    \includegraphics[
        width=\linewidth,
        trim={0cm 4cm 0cm 4cm},
        clip
    ]{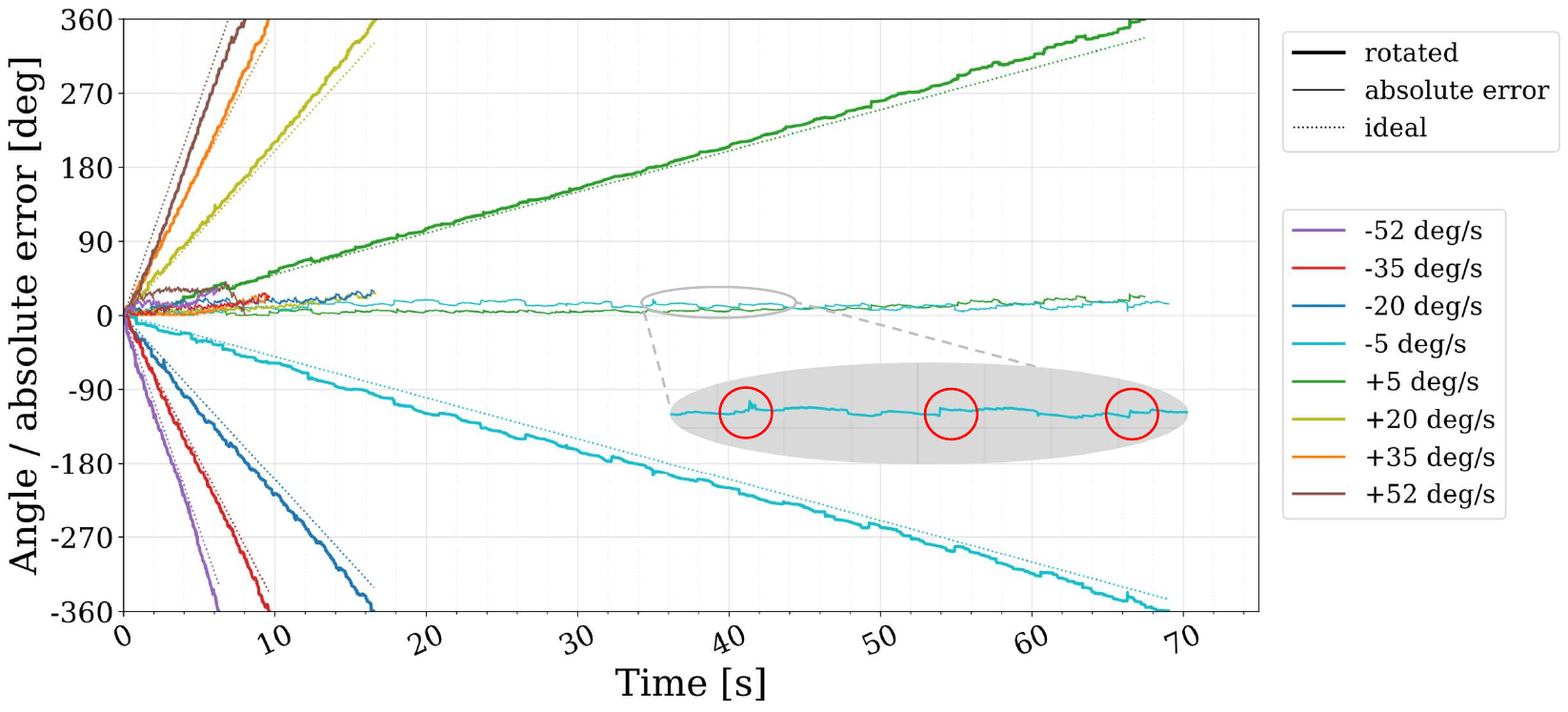}
    \caption{
        Catheter rotation at tested positive and negative angular velocities. Solid lines show measured rotations, dashed lines ideal profiles, and thin lines absolute error. Red circles indicate leader--follower switching
    }
    \label{fig:cath_rot}
\end{figure}

Fig.~\ref{fig:gw_trans}--~\ref{fig:cath_rot} show the guidewire and catheter manipulation experiments. Fig.~\ref{fig:gw_trans} and Fig.~\ref{fig:cath_trans} present traveled distance and absolute error for all insertion and retraction velocities, whereas Fig.~\ref{fig:gw_rot} and Fig.~\ref{fig:cath_rot} show rotational displacement and corresponding error for all rotational velocities. Solid lines show measured motion, dashed lines ideal profiles, and thin lines absolute error. Red circles indicate leader-follower switching.

\begin{table}[t]
\centering
\caption{Mean absolute errors for all guidewire and catheter translational and rotational experiments}
\label{tab:MAE_trans_rot}
\begin{tabular}{c|c|c}
\hline
Velocity & Guidewire & Catheter \\
\hline

[\si{\milli\meter\per\second}]
& \multicolumn{2}{c}{Translation} \\
\hline

-19 & $5.6 \pm 3.7$ & $2.7 \pm 1.8$ \\
-15 & $3.1 \pm 2.0$ & $3.5 \pm 2.3$ \\
-10 & $10.6 \pm 6.2$ & $9.4 \pm 5.5$ \\
-5  & $10.6 \pm 6.1$ & $12.3 \pm 7.6$ \\
5   & $13.2 \pm 7.8$ & $15.6 \pm 8.4$ \\
10  & $4.4 \pm 3.4$ & $3.4 \pm 2.0$ \\
15  & $2.2 \pm 0.8$ & $11.7 \pm 4.1$ \\
19  & $3.0 \pm 1.2$ & $4.0 \pm 2.8$ \\

\hline

[\si{\degree\per\second}]
& \multicolumn{2}{c}{Rotation} \\
\hline

-52 & $16.0 \pm 6.7$ & $16.0 \pm 6.7$ \\
-35 & $13.4 \pm 9.0$ & $10.7 \pm 6.0$ \\
-20 & $15.6 \pm 10.8$ & $16.4 \pm 5.7$ \\
-5  & $19.6 \pm 13.4$ & $11.6 \pm 4.0$ \\
5   & $18.6 \pm 10.8$ & $7.7 \pm 5.4$ \\
20  & $18.6 \pm 10.5$ & $10.8 \pm 6.9$ \\
35  & $19.1 \pm 10.1$ & $7.2 \pm 7.0$ \\
52  & $4.3 \pm 3.3$ & $16.0 \pm 6.7$ \\

\hline
\end{tabular}
\end{table}

Table~\ref{tab:MAE_trans_rot} summarizes the mean absolute errors for each tested velocity and instrument across all runs. The overall MAE across all translational runs is \SI{7.2 \pm 4.4}{\milli\meter} and across all rotational runs \SI{14.6 \pm 5.6}{\degree}.
For the guidewire translational experiments (Fig.~\ref{fig:gw_trans}), the measured insertion and retraction motions exhibit a strong linear correlation with the commanded trajectories across all tested velocities.
The highest mean absolute error was observed at a velocity of \SI{5}{\milli\meter\per\second}, reaching \SI{13.2}{\milli\meter}, whereas the lowest mean absolute error of \SI{2.2}{\milli\meter} occurred at \SI{15}{\milli\meter\per\second}.
 
Similar results are observed for catheter translation in Fig.~\ref{fig:cath_trans}.
The highest mean absolute error was observed at a velocity of \SI{5}{\milli\meter\per\second}, reaching \SI{15.6}{\milli\meter}, whereas the lowest mean absolute error of \SI{2.7}{\milli\meter} occurred at \SI{-19}{\milli\meter\per\second}.

Catheter manipulation exhibits a smoother motion profile than the guidewire experiments, possibly because the higher inertia of the catheter reduces oscillatory motion, particularly during leader-follower switching events. Error also tends to increase with decreasing velocity, possibly because lower velocities require more time to reach the target displacement of \SI{200}{\milli\meter} and therefore allow small tracking errors to accumulate. This is supported by smaller tracking errors at short traveled distances and larger errors at larger displacements.

The guidewire rotational experiments in Fig.~\ref{fig:gw_rot} show a strong correlation between measured rotational displacement and the ideal trajectories. However, linearity decreases after approximately \SI{120}{\degree} of rotation.
The highest mean absolute error of \SI{19.6}{\degree} was observed at an angular velocity of \SI{-5}{\degree\per\second}, whereas the lowest mean absolute error of \SI{4.3}{\degree} occurred at \SI{52}{\degree\per\second}.

The rotational movement of the catheter shown in Fig.~\ref{fig:cath_rot} remains consistent throughout the experiments and closely follows the commanded trajectory.
The largest mean absolute error of \SI{16.4}{\degree} was recorded at \SI{-20}{\degree\per\second}, whereas the most accurate tracking performance, with a mean absolute error of \SI{7.2}{\degree}, was achieved at \SI{35}{\degree\per\second}.
As in the translational motion profiles, short spikes are also observed in the rotational error curves and are attributed to leader-follower handover.

\subsection{In Vitro Phantom Experiment}
All attempts to navigate the guidewire to the aortic root and vertebral artery were successful during manual navigation. In contrast, one robot-assisted attempt to reach the vertebral artery was unsuccessful, whereas all other robot-assisted attempts were completed successfully. This resulted in a success rate of 100\% for manual navigation and 90\% for robot-assisted navigation.

\begin{figure}[htbp]
    \centering
    
    \includegraphics[width=\linewidth,
    clip]{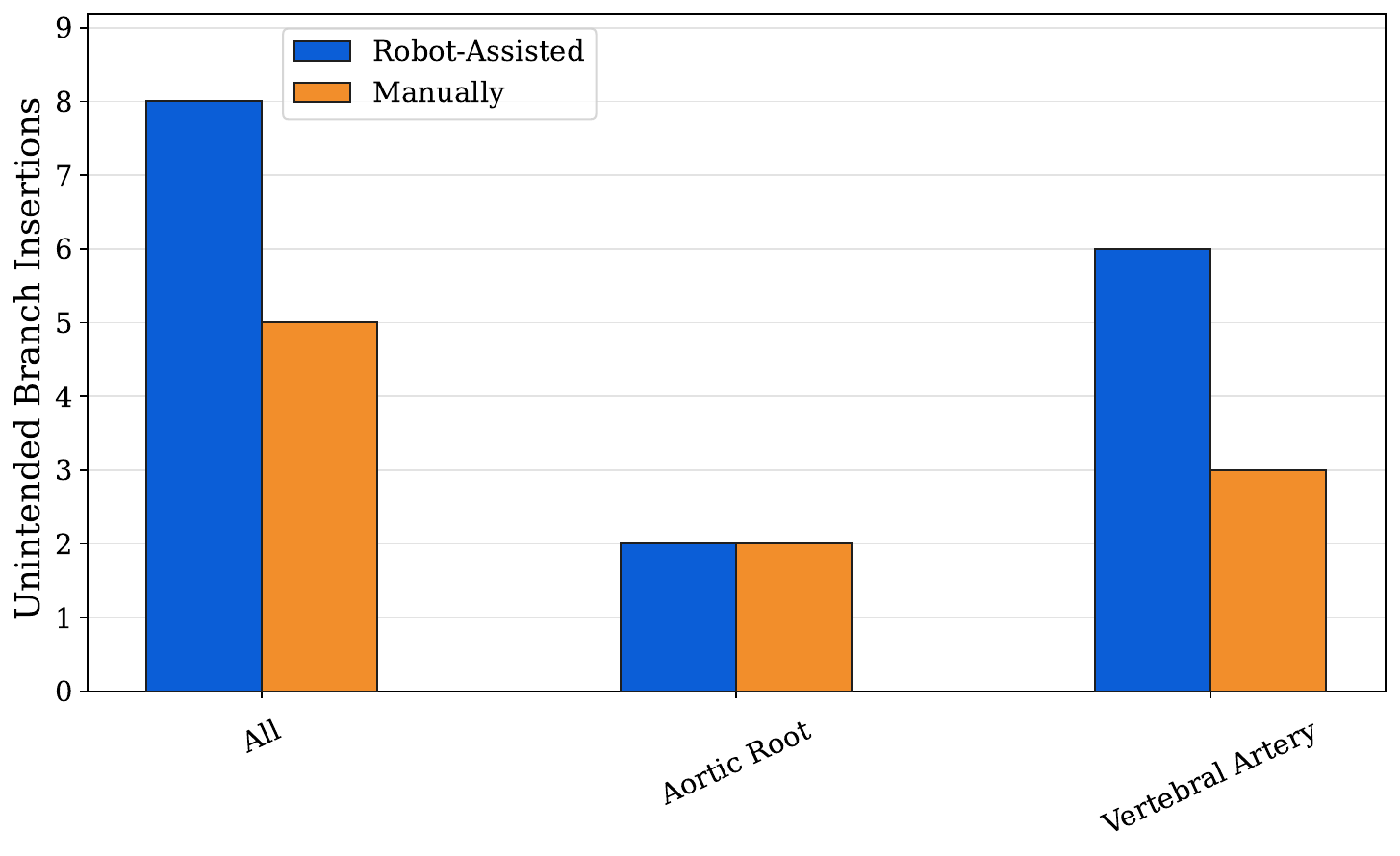}
    
    \vspace{0.5em}

    \caption{Unintended branch insertions for robotic (blue) and manual operation (orange) in all trials, aortic-root and vertebral-artery navigation}
    
    \label{fig:in_vitro_plot_failures}
\end{figure}

Fig.~\ref{fig:in_vitro_plot_failures} shows the number of unintended branch insertions for manual navigation (orange) and robot-assisted navigation (blue). Fewer occurred during manual navigation (5) than during robot-assisted navigation (8). This difference is attributable to the vertebral artery target, whereas the number was identical for the aortic root.

\begin{figure}[htbp]
    \centering
    
    \includegraphics[width=\linewidth,
    clip]{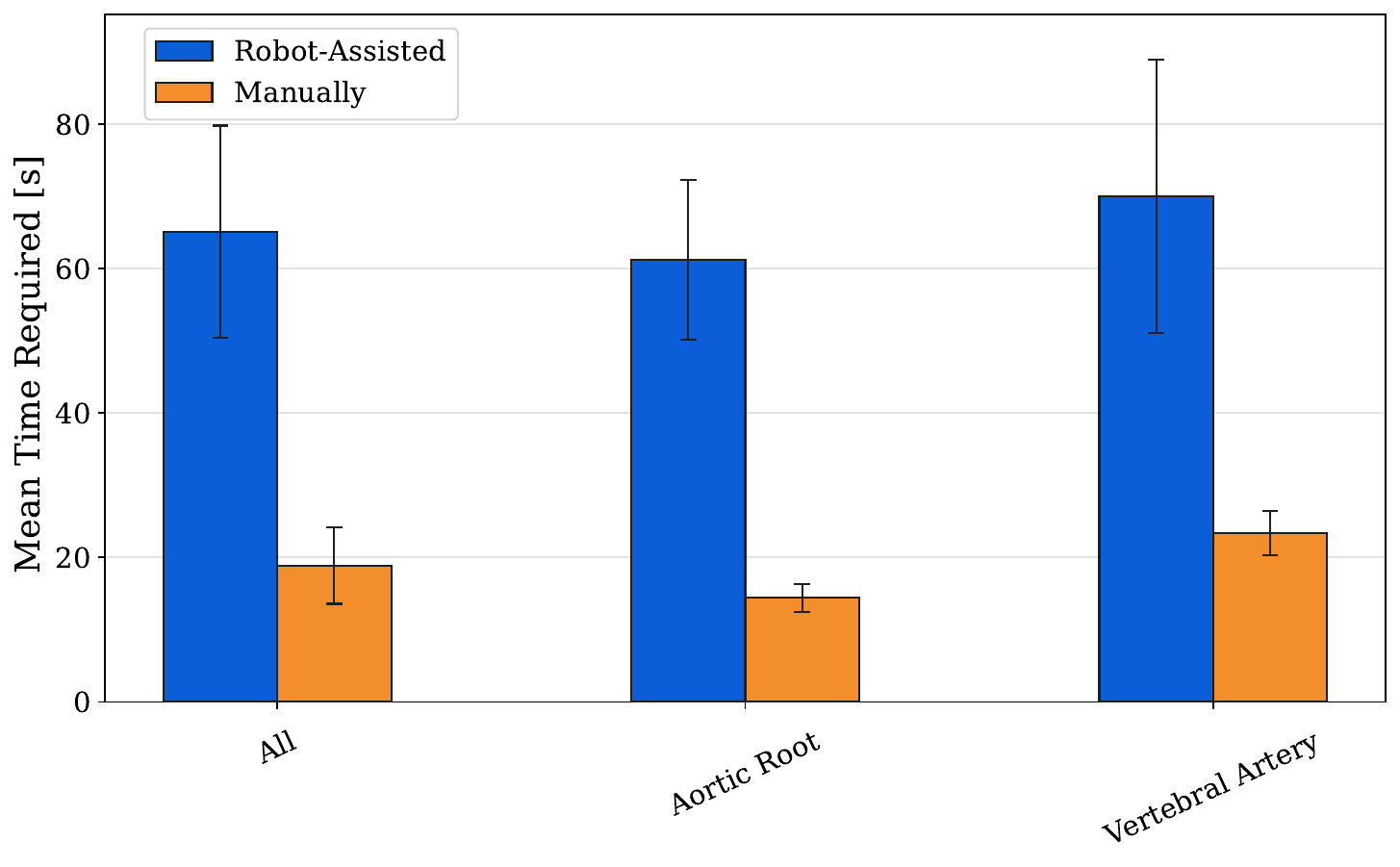}
    
    \vspace{0.5em}

    \caption{Mean time for robotic (blue) and manual operation (orange) in all trials, aortic-root and vertebral-artery navigation}
    
    \label{fig:in_vitro_plot_mean_time}
\end{figure}

Fig.~\ref{fig:in_vitro_plot_mean_time} shows the mean time required to reach a target from the starting position for manual navigation (orange) and robot-assisted navigation (blue). Manual navigation outperformed robot-assisted navigation for both target locations. The overall mean time was \SI{19}{\second} for manual navigation and \SI{65}{\second} for robot-assisted navigation.

\section{Discussion}

This work proposes a new manipulator for endovascular interventions. Its performance was evaluated by quantifying translational and rotational tracking errors and testing the system in a vascular phantom under conditions closer to clinical use.

Overall, the measured motion curves were predominantly smooth and linear. The largest mean relative tracking error for translational motion was \SI{7.8}{\percent}, occurring at \SI{5}{\milli\meter\per\second} during catheter movement. For a preliminary prototype evaluation, this indicates controlled translational motion, although task-specific clinical accuracy requirements remain to be defined. For larger travel distances, operators typically use velocities higher than \SI{5}{\milli\meter\per\second}; at those velocities, translational tracking errors ranged from \SI{1.1}{\percent} to \SI{5.9}{\percent}. The largest mean relative rotational tracking error was \SI{5.4}{\percent}, occurring at \SI{-5}{\degree\per\second} during guidewire movement. These results suggest feasible rotational actuation under bench-top conditions, but further validation is required to determine whether it is sufficient for clinical navigation tasks.

The handover between the two carts was visible in the measured motion curves. Although its effect remained limited to the single-digit millimeter and degree range, errors increased at slower velocities, where longer motion duration allowed deviations to accumulate. These errors could be further reduced by integrating an instrument-tracking module that measures actual instrument motion in real time and enables closed-loop correction.

In the vascular phantom experiments, the success rate of \SI{90}{\percent} provides preliminary evidence that the proposed concept can support endovascular navigation tasks in an in vitro setting. The failed attempt can be attributed to insufficient guidewire rotation caused by entanglement of the guidewire tip within the silicone vessel tubes and insufficient torque transmission from the gripper to the guidewire. The longer completion times observed during robot-assisted trials can be attributed to the limited maximum velocity of the robot, \SI{19}{\milli\meter\per\second}. In contrast, the average velocities during manual manipulation were \SI{48.3}{\milli\meter\per\second} for navigation to the aortic root and \SI{33.7}{\milli\meter\per\second} for navigation to the vertebral artery. The practical velocity limit of the manipulator was lower than the analytical upper bound because the theoretical estimate considers only idealized clamping and return times and not the discrete implementation of the handover sequence. In the implementation, the control-loop time averaged \SI{13}{\milli\second}. A leader-follower handover is not completed within a single control-loop iteration, but requires several sequential state transitions, including detection of the leader stroke limit, acceleration and velocity verification of the follower, initiation of clamping, verification of the completed clamp after the \SI{50}{\milli\second} clamping delay, and exchange of the leader and follower roles. These delays reduce the effective time available for continuous instrument motion and thus the practical maximum velocity compared with the analytical maximum.

The phantom experiments were conducted by a single non-clinical user, hence the resulting performance measures should be interpreted with caution. More reliable conclusions regarding clinical readiness require studies with multiple users, including experienced surgeons. Future evaluations should also consider in vivo experiments and surgeon feedback.

The presented concept can manipulate different endovascular instruments, such as guidewires and catheters, without changing the robotic setup. This is enabled by the pneumatically actuated gripper and gripper-gear design, which allows rapid instrument exchange. Such functionality is relevant for endovascular procedures, in which multiple instruments are commonly used during a single intervention. Compared with systems for which platform setup, positioning, or instrument exchange can contribute to longer procedure times, the top-loading design of the robot may simplify instrument handling and reduce workflow interruptions during instrument changes \cite{dagnino_-vivo_2023}. Furthermore, the compact patient-side design addresses another limitation reported for larger robotic setups, namely operating-room space requirements \cite{song_novel_2023}. Together, rapid instrument exchange and a compact patient-side footprint address key workflow requirements.

In addition, the modular two-cart design may facilitate sterilization or enable disposable components for parts that are difficult to sterilize. In future setups, the compact architecture could allow multiple manipulators to be arranged in series while maintaining a small patient-side footprint.

\section{Conclusion}

This work presents a compact robotic system for endovascular interventions based on two alternating carts with pneumatically actuated membrane grippers. The top-loading gripper design enables rapid exchange of endovascular instruments without changing the robotic setup, while the leader-follower control strategy allows continuous translational and rotational actuation despite the limited stroke of each cart. Experimental evaluation demonstrates guidewire and catheter manipulation with predominantly smooth translational and rotational motion under bench-top conditions.

In vitro phantom experiments further show the feasibility of robot-assisted navigation in a vascular model with pulsatile flow. Although manual navigation was faster and resulted in fewer unintended branch insertions, the manipulator reached the target in most trials. The observed limitations, including insufficient guidewire rotation in one failed trial and the limited practical maximum velocity, indicate that future iterations should improve torque transmission and reliable actuation speed.

Future work will focus on increasing gripping performance, improving rotational reliability, and integrating force sensing for haptic feedback. In addition, the system should be evaluated in studies with multiple users, including experienced surgeons, and in more clinically realistic settings to further assess clinical usability and workflow benefits. Future work will also explore simultaneous manipulation of guidewires and catheters using two robot units.

\backmatter

\section*{Declarations}

\noindent\textbf{Funding} This work was supported by the state of Bavaria through Bayerische Forschungsstiftung (BFS) under Grant AZ-1633-24-TED-MeD.

\vspace{0.5em}
\noindent\textbf{Competing Interests} The authors declare that they have no conflict of interest.

\vspace{0.5em}
\noindent\textbf{Data availability} Data are available from the authors on reasonable request.

\vspace{0.5em}
\noindent\textbf{Author contributions} Jonas Fischer and Lennart Karstensen contributed equally to this work.

\bibliography{references}

\end{document}